\documentclass{article}

\usepackage{microtype}
\usepackage{graphicx}
\usepackage{subfigure}
\usepackage{xcolor}
\usepackage{tcolorbox}

\definecolor{hotPink}{RGB}{200, 40, 120}
\definecolor{ObjBlue}{RGB}{36,82,139}
\definecolor{ObjOrange}{RGB}{196,104,0}
\usepackage{booktabs} 

\usepackage{hyperref}

\usepackage[accepted]{icml2025}

\usepackage{amsmath}
\usepackage{amssymb}

\usepackage{mathtools}
\usepackage{amsthm}
\usepackage{graphicx}
\usepackage{subcaption}

\usepackage[capitalize,noabbrev]{cleveref}

\theoremstyle{plain}

\theoremstyle{definition}

\theoremstyle{remark}

\usepackage[textsize=tiny]{todonotes}

\icmltitlerunning{Submission and Formatting Instructions for ICML 2025}

\begin{document}

\twocolumn[
\icmltitle{K-Score: Kalman Filter as a Principled Alternative to \\ Reward Normalization in Reinforcement Learning}



\icmlsetsymbol{equal}{*}


\icmlsetsymbol{equal}{*}

\begin{icmlauthorlist}
\icmlauthor{Zixuan Xia}{equal,bern}
\icmlauthor{Quanxi Li}{equal,bern}
\end{icmlauthorlist}

\icmlaffiliation{bern}{University of Bern, Bern, Switzerland}

\icmlcorrespondingauthor{Zixuan Xia}{zixuan.xia@students.unibe.ch}
\icmlcorrespondingauthor{Quanxi Li}{quanxi.li@students.unibe.ch}

\icmlkeywords{Reinforcement Learning, Reward Normalization, Kalman Filtering}

\vskip 0.3in
]



\icmlkeywords{Machine Learning, ICML}

\vskip 0.3in



\printAffiliationsAndNotice{\icmlEqualContribution} 

\begin{abstract}
We propose a simple yet effective alternative to reward normalization in policy gradient reinforcement learning by integrating a 1D Kalman filter for online reward estimation. Instead of relying on fixed heuristics, our method recursively estimates the latent reward mean, smoothing high-variance returns and adapting to non-stationary environments. This approach incurs minimal overhead and requires no modification to existing policy architectures. Experiments on \textit{LunarLander} and \textit{CartPole} demonstrate that Kalman-filtered rewards significantly accelerate convergence and reduce training variance compared to standard normalization techniques. Code is available at \url{https://github.com/Sumxiaa/Kalman_Normalization}.
\end{abstract}

\section{Introduction}

Reinforcement Learning (RL) has emerged as a powerful paradigm for sequential decision-making, enabling agents to learn optimal behaviors through interaction with the environment. Among the broad family of RL algorithms, \textit{policy gradient methods} stand out for their effectiveness in high-dimensional and continuous action spaces~\cite{sutton1999policy}, where value-based methods often struggle due to discretization challenges or instability in action selection.

Despite their theoretical appeal, policy gradient methods suffer from high variance in gradient estimates, primarily caused by the stochastic nature of both the environment and the policy. High-variance returns can lead to noisy updates, poor sample efficiency, and unstable learning, especially in early-stage training. As a result, numerous variance reduction techniques have been proposed to improve training dynamics and generalization.

One commonly adopted strategy is \textit{reward normalization}, where observed returns are standardized via Z-score normalization using running estimates of the mean and standard deviation~\cite{mnih2016asynchronous, schulman2015trust}. This heuristic smooths fluctuations in reward signals and improves the conditioning of policy gradient updates. However, such techniques are often sensitive to hyperparameters (e.g., decay rates), require careful tuning, and crucially rely on the assumption that reward statistics remain stationary throughout training—a condition rarely satisfied in dynamic or partially observable environments.

To address these limitations, we explore a \textit{principled and adaptive alternative} grounded in Bayesian estimation theory: Kalman filtering~\cite{kalman1960new}. Originally developed for linear state estimation under uncertainty, Kalman filters have found extensive use in robotics, tracking, and control~\cite{welch1995introduction}, but their potential for reward modeling in reinforcement learning remains underexploited.

In this paper, we propose to replace traditional reward normalization with a lightweight, one-dimensional Kalman filter that recursively estimates the latent mean of the return signal. Our approach models the expected reward as a latent stochastic process subject to both process and observation noise. By fusing prior knowledge with new observations in a statistically optimal manner, the Kalman filter provides a dynamic, uncertainty-aware estimate of the reward baseline. This allows the agent to better cope with noisy feedback, non-stationarity, and abrupt shifts in reward distributions.

Our method is simple to implement, computationally efficient, and architecture-agnostic. It requires no change to the policy network, making it compatible with a wide range of existing policy gradient algorithms such as REINFORCE, Actor-Critic, GAE, and PPO. Furthermore, the Kalman filter naturally adapts to the variability of the learning environment without requiring additional hyperparameter tuning beyond initial noise estimates.

We evaluate our method on two well-established control tasks from OpenAI Gym—\textit{CartPole-v1} and \textit{LunarLander-v2}—which respectively represent low-variance dense reward and high-variance sparse reward settings. Experimental results demonstrate that Kalman-filtered reward estimation leads to smoother training, faster convergence, and improved stability compared to conventional Z-score normalization.

\paragraph{Our main contributions are summarized as follows:}
\begin{itemize}
    \item We propose a novel application of Kalman filtering for reward estimation in policy gradient methods, offering a principled replacement for heuristic normalization strategies.
    \item We design a lightweight and efficient 1D Kalman filter module that adaptively tracks non-stationary reward signals in online learning settings.
    \item We empirically validate the effectiveness of our approach across multiple policy gradient algorithms and benchmark environments, showing consistent improvements in convergence speed and stability.
\end{itemize}

\section{Related Work}

\subsection{Policy Gradient Methods}

Policy gradient methods directly optimize a parameterized policy
$\pi_\theta(a|s)$ by estimating gradients of the expected return
\cite{sutton1999policy}. A standard form of the policy gradient is
\begin{equation}
\nabla_\theta J(\theta)
=
\mathbb{E}_{\pi_\theta}
\left[
\nabla_\theta \log \pi_\theta(a_t|s_t)\hat{A}_t
\right],
\end{equation}
where $\hat{A}_t$ denotes an estimate of the advantage. Early methods
such as REINFORCE~\cite{williams1992simple} use Monte Carlo returns
as the learning signal, which yields an unbiased but often high-variance
gradient estimate. Actor-critic methods reduce this variance by learning
a value-function baseline, while Generalized Advantage Estimation
(GAE)~\cite{schulman2015high} further introduces a bias-variance
trade-off through exponentially weighted temporal-difference residuals.
Proximal Policy Optimization (PPO)~\cite{schulman2017proximal}
improves stability by constraining policy updates with a clipped
surrogate objective.

Despite these advances, policy-gradient methods remain sensitive to
the scale and variability of the return or advantage signal. In practice,
normalizing these signals is often crucial for stable training. Our work
focuses on this normalization step and proposes a lightweight Bayesian
filtering mechanism that can be inserted into standard policy-gradient
pipelines without modifying the policy architecture.

\subsection{Reward and Advantage Normalization}

Reward, return, and advantage normalization are widely used to reduce
gradient variance and improve optimization stability in policy-gradient
reinforcement learning. A common approach is Z-score normalization,
which standardizes a return or advantage signal using running estimates
of its mean and standard deviation:
\begin{equation}
\hat{G}_t = \frac{G_t-\mu_t}{\sigma_t+\epsilon}.
\end{equation}
Although simple and effective, such running-statistics methods are
typically heuristic. Their behavior depends on decay rates and window
sizes, and they do not explicitly model uncertainty in the estimated
statistics. This becomes problematic when the reward distribution changes
during learning, as is common in online RL where the policy itself is
non-stationary.

Several approaches address related issues of scale and stability. PopArt
\cite{van2018deep} rescales value-function targets while preserving the
un-normalized output, enabling learning across tasks with different reward
magnitudes. Multi-task reinforcement learning methods have also used
normalization to handle heterogeneous reward scales across environments
\cite{hessel2019multi}. Meta-gradient methods~\cite{xu2018meta} adapt
learning hyperparameters online, including quantities related to learning
signals, but usually introduce additional computational and algorithmic
complexity.

Our method takes a simpler probabilistic view: the observed return or
advantage is treated as a noisy observation of a latent learning signal.
Rather than maintaining heuristic running moments, we estimate this latent
signal and its uncertainty online with a scalar Kalman filter. This yields
an adaptive normalization rule that remains plug-and-play for standard
policy-gradient algorithms.

\subsection{Kalman Filtering in Reinforcement Learning}

Kalman filtering is a classical Bayesian method for online state
estimation under process and observation noise
\cite{kalman1960new,welch1995introduction}. In reinforcement learning,
Kalman-style methods have mainly been explored for value estimation,
uncertainty modeling, or belief-state inference. For example, Shashua and
Mannor~\cite{shashua2019trust} use Kalman filtering to track uncertainty
in value-function approximation for trust-region policy optimization.
Painter-Wakefield and Parr~\cite{painter2012greedy} study sparse
reinforcement learning methods that use filtering-style estimation for
local value modeling. Kalman filters and related Bayesian filters are also
natural tools in partially observable settings, where the agent must infer
hidden states from noisy observations.

Most of these works use filtering to estimate states, values, or model
uncertainty. In contrast, our method applies a lightweight one-dimensional
Kalman filter directly to the reward, return, or advantage normalization
signal. The goal is not to replace the value function or modify the policy
update rule, but to provide an uncertainty-aware alternative to conventional
running mean and standard-deviation normalization.

\subsection{Concurrent Work}

A closely related concurrent work is KRPO~\cite{wang2025kalman}, which
introduces Kalman filtering into reinforcement learning for language-model reasoning. KRPO replaces the group-mean baseline in GRPO with a Kalman-filtered estimate of the reward mean and variance, thereby improving the stability of advantage estimation in LLM post-training. {\color{hotPink}Our work differs in scope and setting. KRPO focuses on GRPO-style reinforcement learning for language-model reasoning, where rewards are computed over groups of sampled completions. In contrast, we study Kalman-based normalization as a general return or advantage preprocessing mechanism for policy-gradient methods in classic control environments,
including REINFORCE-style updates, actor-critic methods, GAE, and PPO. Thus, the two works are complementary: KRPO demonstrates the usefulness of Kalman filtering in LLM reasoning with group-relative objectives, while
our work investigates the same filtering principle as a general normalization module for policy-gradient reinforcement learning.}

\section{Methods}

\subsection{Problem Formulation and Motivation}

In policy gradient methods, the quality and consistency of the reward signal are critical for stable and efficient learning. However, in many environments—particularly those with sparse or noisy feedback—the reward signal can exhibit high variance, causing unstable updates and slow convergence.

To mitigate this, practitioners often employ reward normalization techniques. A widely used approach is Z-score normalization, which standardizes returns using running estimates of the mean and standard deviation. Despite its empirical success, this method has several well-known limitations:
\begin{itemize}
    \item It assumes a stationary reward distribution, which rarely holds in practice.
    \item It requires careful tuning of decay parameters for moving averages.
    \item It provides no formal treatment of uncertainty or noise in the reward signal.
\end{itemize}

These issues motivate the need for a more principled, adaptive, and uncertainty-aware approach to reward estimation and normalization. Inspired by Bayesian filtering, we propose to model the true underlying reward signal as a latent stochastic process. Instead of assuming static reward statistics, we treat the reward mean as a hidden variable that evolves over time and must be estimated from noisy observations.

This leads us naturally to the Kalman filter, a recursive estimator that fuses prior predictions and observed data in a statistically optimal manner. In the next section, we describe how we adapt this framework to the reinforcement learning setting.

\subsection{Kalman Filtering for Adaptive Reward Estimation}

We model the return signal $G_t$ as a noisy observation of a latent reward mean $x_t$, which evolves over time according to a first-order Markov process. Specifically, we define the process model and observation model as:

\textbf{Latent Reward Dynamics:}
\begin{align}
x_t &= x_{t-1} + w_t, \quad w_t \sim \mathcal{N}(0, Q), \\
G_t &= x_t + v_t, \quad v_t \sim \mathcal{N}(0, R),
\end{align}
where $w_t$ and $v_t$ represent process noise and observation noise, respectively, and $Q$, $R$ are their variances.

Given these dynamics, the Kalman filter maintains a recursive estimate of the reward mean $x_t$ and its uncertainty $P_t$ via a two-step update:

\textbf{Predict Step:}
\begin{align}
x_{t|t-1} &= x_{t-1}, \\
P_{t|t-1} &= P_{t-1} + Q,
\end{align}

\textbf{Update Step:}
\begin{align}
K_t &= \frac{P_{t|t-1}}{P_{t|t-1} + R}, \\
x_t &= x_{t|t-1} + K_t (G_t - x_{t|t-1}), \\
P_t &= (1 - K_t) P_{t|t-1}.
\end{align}

Using the filtered estimate $x_t$ and variance $P_t$, we define the normalized return:
\begin{equation}
\hat{G}_t^{\text{Kalman}} = \frac{G_t - x_t}{\sqrt{P_t + \epsilon}},
\end{equation}
where $\epsilon$ is a small constant to prevent division by zero.

This formulation provides a dynamically updated baseline that automatically adapts to changes in the reward signal and incorporates uncertainty into the normalization process. Unlike Z-score normalization, which applies uniform decay over time, the Kalman gain $K_t$ dynamically balances prior and observed information based on noise levels.

In practice, this leads to a more robust and data-efficient normalization strategy, especially in the early stages of training or in highly non-stationary environments.

\subsection{Integration into Policy Gradient Algorithms}

We embed the Kalman filtering mechanism into the policy gradient pipeline at the stage where returns or advantages are typically normalized. The Kalman-filtered reward estimate serves as an adaptive baseline, reducing variance in gradient estimation while maintaining sensitivity to non-stationary rewards.

The vanilla policy gradient objective is:
\begin{equation}
\nabla_\theta J(\theta) = \mathbb{E}_\pi \left[ \nabla_\theta \log \pi_\theta(a_t | s_t) \cdot G_t \right],
\end{equation}
where $G_t$ is the discounted return. We replace $G_t$ with a normalized variant:
\begin{equation}
\hat{G}_t = \frac{G_t - x_t}{\sqrt{P_t + \epsilon}},
\end{equation}
where $x_t$ and $P_t$ are the filtered mean and variance of $G_t$ computed using the Kalman equations. This yields a more stable signal for gradient estimation.

The procedure is outlined in Algorithm~\ref{alg:kalman_pg}. Note that our method requires no changes to the policy architecture and incurs minimal computational cost, as the Kalman filter operates on scalar values with constant-time updates.

\begin{algorithm}[H]
\caption{Policy Gradient with Kalman-Normalized Returns}
\label{alg:kalman_pg}
\begin{algorithmic}[1]
\STATE \textbf{Initialize:} policy parameters $\theta$, Kalman filter state $(x_0, P_0)$, noise parameters $Q, R$
\FOR{each training iteration}
    \STATE Sample trajectory $\tau = \{(s_t, a_t, r_t)\}_{t=1}^{N}$ using policy $\pi_\theta$
    \STATE Compute discounted returns $G_t = \sum_{k=t}^{N} \gamma^{k - t} r_k$
    \FOR{each timestep $t$ in $\tau$}
        \STATE Update Kalman filter: {\color{hotPink}$(x_t, P_t) \leftarrow \mathcal{K}.\text{update}(G_t)$}
        \STATE Normalize return: $\hat{G}_t = \dfrac{G_t - x_t}{\sqrt{P_t + \epsilon}}$
    \ENDFOR
    \STATE Compute gradient estimate:
    \[
    g = \frac{1}{N} \sum_{t=1}^{N} \nabla_\theta \log \pi_\theta(a_t | s_t) \cdot \hat{G}_t
    \]
    \STATE Update policy: $\theta \gets \theta + \eta \cdot g$
\ENDFOR
\end{algorithmic}
\end{algorithm}

This integration results in improved sample efficiency, smoother training curves, and enhanced adaptability to changing environments—qualities especially valuable in real-world RL deployments.

\subsection{Theoretical Analysis}

We provide a simple analysis to clarify why Kalman-based normalization can be more adaptive than standard running statistics in reinforcement learning. Our goal is not to prove convergence of the full policy-gradient algorithm, but to analyze the behavior of the scalar estimator used to normalize the return or advantage signal.

\paragraph{Running-mean estimation.}

Consider an observed learning signal $G_t$, such as a Monte Carlo return or an advantage estimate. A standard normalization method maintains a running estimate of its mean. In the simplest stationary case, assume
\begin{equation}
G_t = r_{\mathrm{true}} + v_t,
\qquad
v_t \sim \mathcal{N}(0,\sigma^2),
\end{equation}
where $r_{\mathrm{true}}$ is a fixed latent mean. The sample mean after $t$ observations is
\begin{equation}
\bar{G}_t = \frac{1}{t}\sum_{i=1}^{t} G_i,
\end{equation}
and its mean squared error is
\begin{equation}
\mathbb{E}\left[(\bar{G}_t-r_{\mathrm{true}})^2\right]
=
\frac{\sigma^2}{t}.
\end{equation}
Thus, the sample mean is consistent under stationarity. However, this property also implies that old observations are never forgotten. In online reinforcement learning, the distribution of returns changes as the policy improves, so a purely averaging-based estimator can lag behind the current reward scale.

\paragraph{Kalman filtering as adaptive mean estimation.}

We instead model the latent mean of the learning signal as a slowly evolving process:
\begin{align}
x_t &= x_{t-1} + w_t,
\qquad
w_t \sim \mathcal{N}(0,Q), \\
G_t &= x_t + v_t,
\qquad
v_t \sim \mathcal{N}(0,R),
\end{align}
where $Q$ controls how quickly the latent mean is allowed to drift and $R$ represents observation noise. The scalar Kalman filter updates its estimate as
\begin{equation}
x_t = x_{t|t-1} + K_t(G_t-x_{t|t-1}),
\end{equation}
with Kalman gain
\begin{equation}
K_t = \frac{P_{t|t-1}}{P_{t|t-1}+R}.
\end{equation}
Here $P_{t|t-1}$ is the predicted uncertainty of the latent mean. The gain $K_t$ determines how much the estimator trusts the new observation relative to the previous estimate. When uncertainty is high, $K_t$ is large and the estimator adapts quickly. When observations are noisy, $K_t$ becomes smaller and the estimator smooths the signal more strongly.

\paragraph{Steady-state behavior.}

For the scalar random-walk model, the posterior variance converges to the steady-state solution
\begin{equation}
P_\infty
=
\frac{\sqrt{Q^2+4QR}-Q}{2}.
\end{equation}
When $Q=0$, the latent mean is assumed constant and the Kalman filter reduces to a recursive averaging estimator. When $Q>0$, the filter maintains a non-zero steady-state uncertainty, which prevents the estimator from becoming overconfident and allows it to track non-stationary changes in the return distribution. In the regime $Q \ll R$, this uncertainty scales approximately as
\begin{equation}
P_\infty \approx \sqrt{QR}.
\end{equation}

\paragraph{Interpretation for reinforcement learning.}

This behavior is useful in policy-gradient reinforcement learning because the return or advantage distribution is typically non-stationary. Early in training, the policy changes rapidly and rewards are often high variance. Later in training, the reward scale may shift as the agent discovers better behaviors. Standard running normalization treats these changes only through heuristic decay rates, whereas the Kalman filter provides an explicit mechanism for balancing adaptation and smoothing through the uncertainty ratio between $Q$ and $R$.

The resulting normalized signal is
\begin{equation}
\hat{G}_t^{\mathrm{Kalman}}
=
\frac{G_t-x_t}{\sqrt{P_t+\epsilon}},
\end{equation}
where $x_t$ acts as an adaptive baseline and $P_t$ provides an uncertainty-aware scale estimate. This can reduce the impact of noisy outliers while still allowing the normalization statistics to follow changes in the learning process.

\begin{tcolorbox}[colback=blue!5!white, colframe=blue!75!black, title=\textbf{Advantages of Kalman-based normalization}]

\begin{itemize}
    \item it recovers standard recursive averaging as a special case when the latent mean is stationary;
    \item it introduces controlled forgetting through the process noise $Q$, enabling adaptation to non-stationary return distributions;
    \item it provides an uncertainty-aware gain $K_t$ that automatically balances smoothing and responsiveness.
\end{itemize}
\end{tcolorbox}
These properties explain why Kalman-based normalization can improve stability and sample efficiency in the empirical settings studied in this paper.

\section{Experiment}
\subsection{Experimental Setup}

We evaluate the effectiveness of our Kalman-based reward normalization across two classic control environments from OpenAI Gym:

\begin{itemize}
    \item \textbf{CartPole-v1}: A simple balancing task with a discrete action space and dense reward. The agent receives $+1$ for every timestep it balances the pole, up to a maximum episode length of 500. The environment is considered solved when {\color{ObjBlue}the average reward over 100 episodes exceeds 475.}
    \item \textbf{LunarLander-v2}: A more challenging control task with a larger discrete action space and sparse, high-variance rewards. The task is considered solved when {\color{ObjOrange}the agent achieves an average reward above 200.}
\end{itemize}

All algorithms are implemented using PyTorch and trained using the standard OpenAI Gym interface. Each model is trained for a fixed number of episodes or until it reaches the predefined reward threshold. Due to time constraints and in order to reduce randomness to some content, we set up both a training environment and a separate evaluation environment for each baseline. These environments share identical configurations but are initialized with different random seeds. Training is terminated once the agent reaches the predefined \texttt{REWARD\_THRESHOLD} in the evaluation environment.

For our Kalman reward normalization, we implement two variants:
\begin{itemize}
    \item \textbf{Simple Kalman}: uses fixed process and measurement noise values $(Q, R)$ selected via grid search for each model and environment.
    \item \textbf{Adaptive Kalman}: adaptively updates the measurement noise $R_t$ using an exponential moving average of the squared residuals, with update rule
    \begin{equation}
    R_t = \alpha R_{t-1} + (1 - \alpha)(z_t - x_{t-1})^2
    \end{equation}
    where $\alpha = 0.9$ unless otherwise specified.
\end{itemize}
These variants are used in both baseline comparisons and ablation studies to evaluate the impact of dynamic versus fixed noise modeling.

\subsection{Baselines and Variants}

To evaluate the effect of Kalman-based reward normalization, we compare its performance against several widely used policy gradient methods with standard return or advantage estimation techniques. The baselines include:

\begin{itemize}
    \item \textbf{Actor-Critic (AC)}: A standard policy gradient method where the advantage is computed as $A_t = G_t - V(s_t)$ using a learned value function as a baseline.
    \item \textbf{Generalized Advantage Estimation (GAE)} \cite{schulman2015high}: Computes a bias-variance tradeoff approximation of the advantage by using an exponentially weighted sum of temporal differences.
    \item \textbf{Proximal Policy Optimization (PPO)} \cite{schulman2017proximal}: A robust on-policy optimization method using clipped surrogate objectives and value function baselines.
\end{itemize}

We augment each of these methods with our proposed Kalman-based return normalization. Specifically, we apply a 1D Kalman filter to the computed return $G_t$ before it is used in policy gradient or critic updates. This gives rise to two variants:

\begin{itemize}
    \item \textbf{Simple Kalman}: Uses a fixed noise Kalman filter for reward normalization, with $(Q, R)$ selected via grid search.
    \item \textbf{Adaptive Kalman}: Updates the measurement noise $R_t$ dynamically using an exponential moving average of residuals, as described in Section 4.1.
\end{itemize}

These variants allow us to examine the effect of reward normalization independently of the underlying optimization algorithm, and to study the benefit of adaptive uncertainty modeling in high-variance environments.

\subsection{Evaluation Metric}

To evaluate and compare convergence speed across different algorithms and normalization strategies, we adopt a threshold-based metric. For each environment, we define a fixed \texttt{REWARD\_THRESHOLD} and measure the number of training episodes required for the agent to reach or exceed this threshold in the evaluation environment.

\begin{itemize}
    \item \textbf{CartPole-v1}: \texttt{REWARD\_THRESHOLD} is set to 475, consistent with the standard OpenAI Gym success criterion.
    \item \textbf{LunarLander-v2}: \texttt{REWARD\_THRESHOLD} is set to 200, indicating successful landing behavior over multiple trials.
\end{itemize}

For each method, we record the number of episodes needed until the evaluation environment achieves an average return above the threshold. As described in Section~4.1, training is terminated as soon as this condition is met. This setup reflects a practical goal in real-world RL applications: reaching reliable performance with minimal training steps.

The reported metric is the episodes to convergence, which can represent the learning efficiency, rather than final asymptotic performance, and highlights the potential of Kalman-based reward normalization to accelerate policy optimization.

\begin{table}[t]
\centering
\caption{Convergence speed of PPO with standard and Kalman-based reward normalization. We report the number of episodes required to reach the target reward. Lower is better.}
\label{tab:ppo_convergence}
\begin{tabular}{lcccc}
\toprule
Env  & Z-Score & K-Score (Ours) & Imp \\
\midrule
CartPole-v1 & 306 & \textbf{77} & \textcolor{green!45!black}{$3.97\times$} \\
LunarLander-v2  & 991 & \textbf{750} & \textcolor{green!45!black}{$1.32\times$} \\
\bottomrule
\end{tabular}
\end{table}
\subsection{Results and Analysis}

We evaluate our method in terms of convergence speed, measured by the number of episodes required to reach a predefined reward threshold in the evaluation environment. All experiments are conducted using separate training and evaluation environments with different random seeds to assess generalization performance.

\begin{figure}[H]
    \centering
    \includegraphics[width=0.9\linewidth]{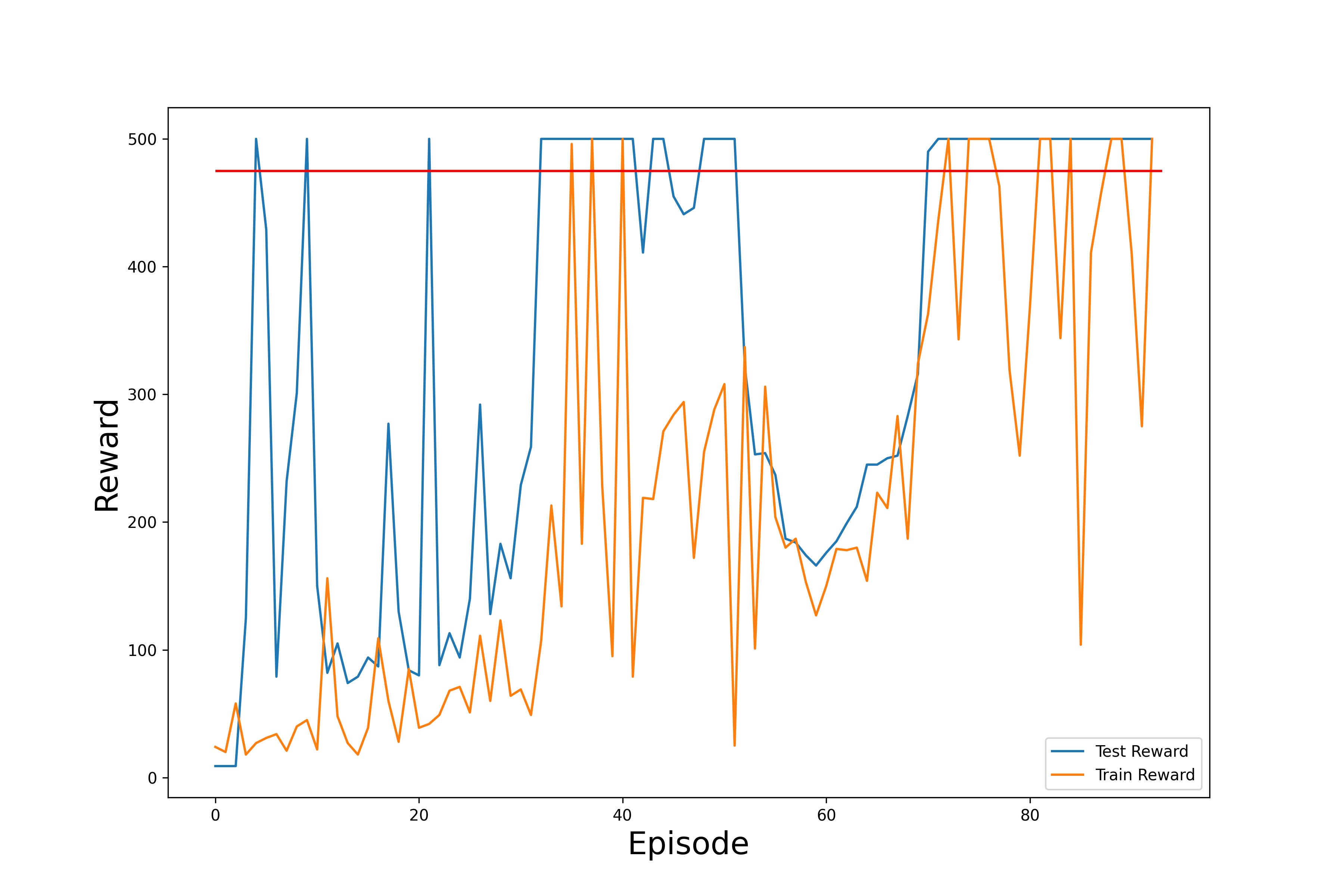}
    \caption{Actor-Critic on CartPole using Z-score normalization. High variance in early training causes slow convergence.}
    \label{fig:cartpole_ac_zscore}
\end{figure}

\begin{figure}[H]
    \centering
    \includegraphics[width=0.9\linewidth]{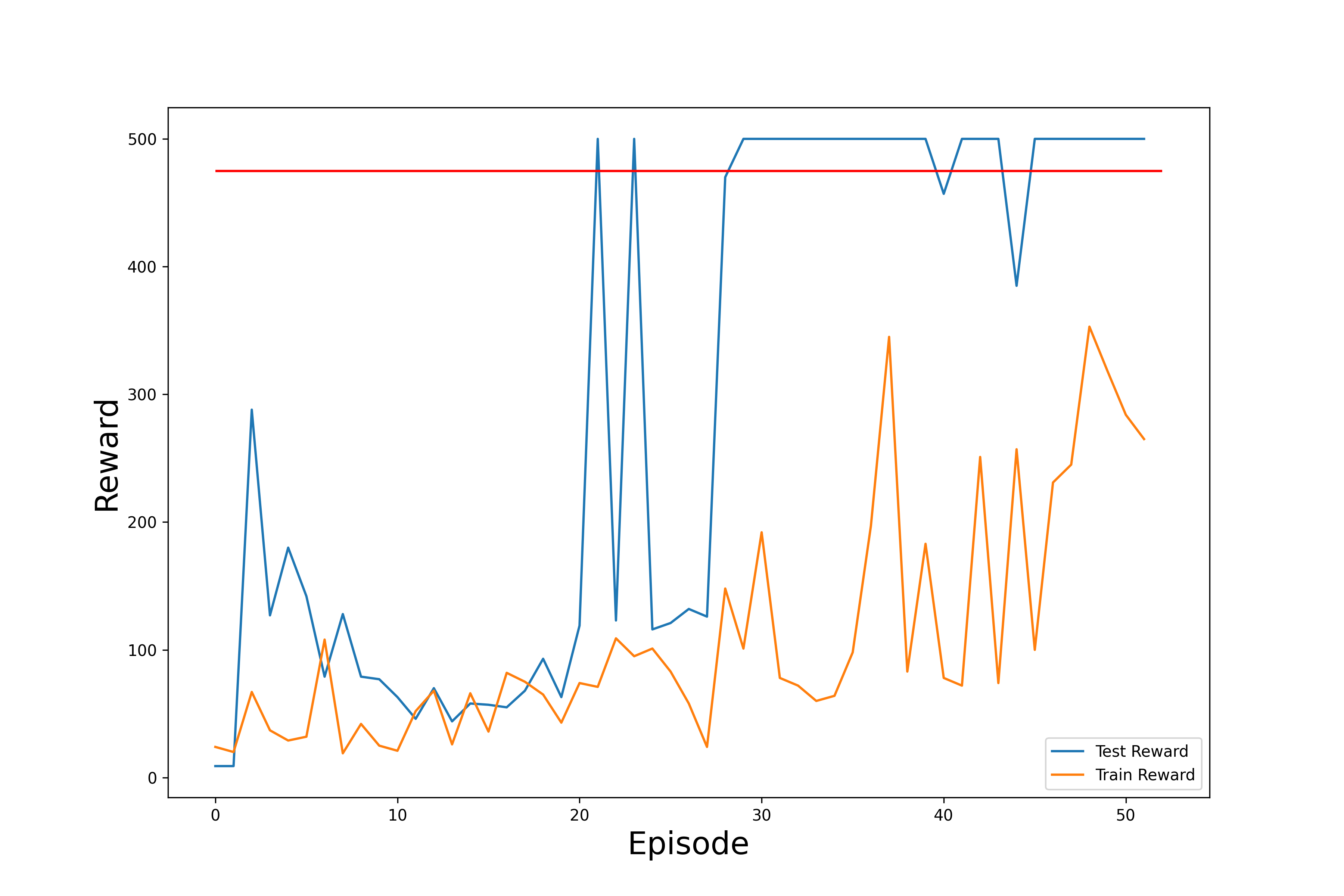}
    \caption{Actor-Critic on CartPole using Kalman normalization. Faster and more stable convergence compared to Z-score.}
    \label{fig:cartpole_ac_kalman}
\end{figure}

\paragraph{CartPole with Actor-Critic}
Figure~\ref{fig:cartpole_ac_zscore} and Figure~\ref{fig:cartpole_ac_kalman} compare the training performance of the Actor-Critic algorithm on the CartPole environment using two different reward normalization methods: Z-score normalization and Kalman-based normalization. Under Z-score normalization (Figure~\ref{fig:cartpole_ac_zscore}), the policy eventually reaches the target reward threshold of 475, but only after approximately 70 episodes. During the early phase of training (episodes 0--50), both the training and evaluation rewards fluctuate significantly, with frequent spikes and collapses, indicating unstable learning dynamics and sensitivity to noise in the return signal.

In contrast, Kalman-based normalization (Figure~\ref{fig:cartpole_ac_kalman}) enables the agent to achieve the reward threshold much earlier—within 25 episodes—and maintain consistent performance thereafter. The training curve is noticeably smoother, and the test reward stabilizes around the optimal level with minimal variance. This improvement can be attributed to the Kalman filter’s ability to incorporate uncertainty into its reward estimation, allowing the policy to ignore early outliers and adapt cautiously in high-variance regimes. These results highlight the effectiveness of Kalman normalization in accelerating convergence and enhancing stability, especially during the critical early stages of training.

\paragraph{CartPole with GAE}

\begin{figure}[H]
    \centering
    \includegraphics[width=0.8\linewidth]{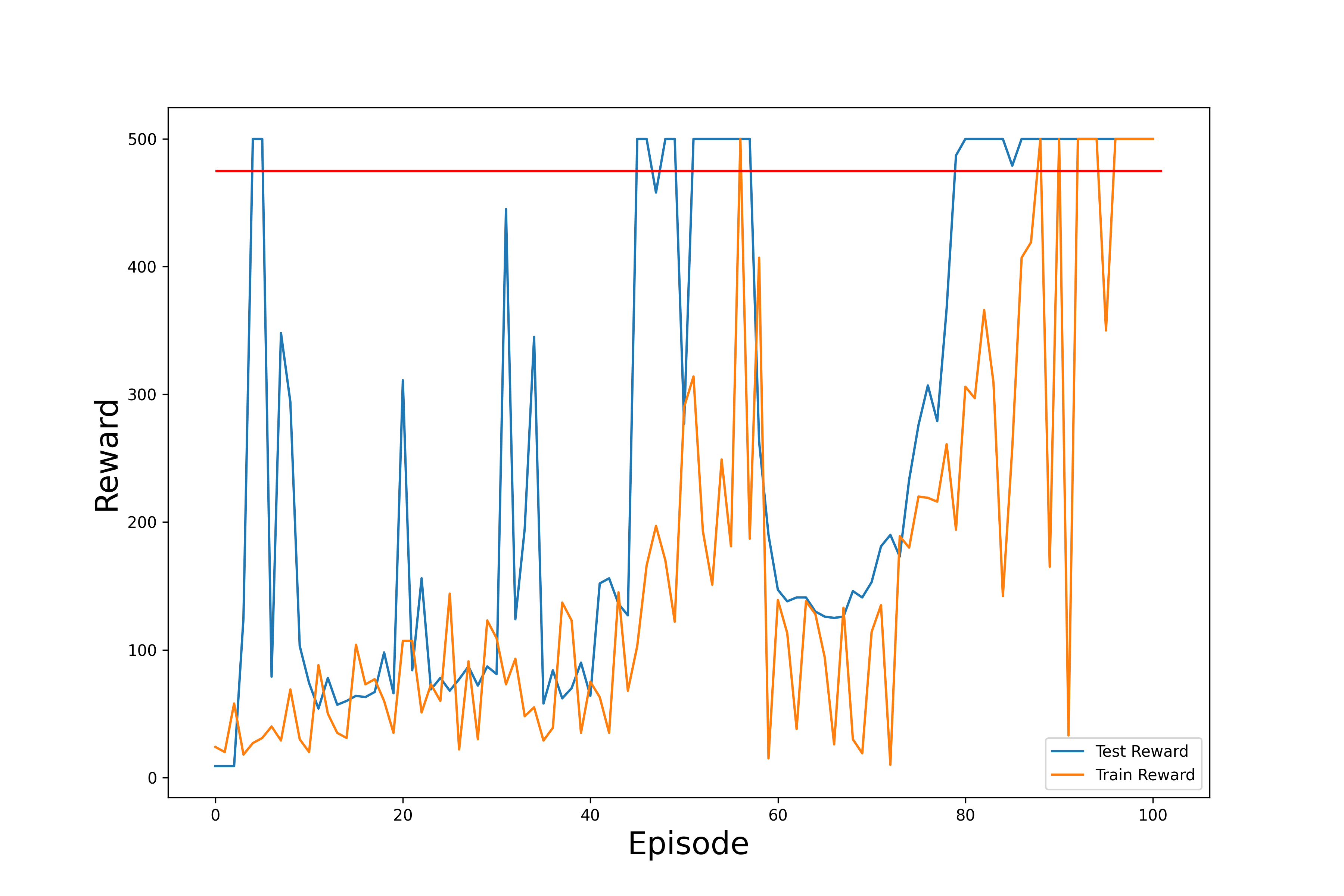}
    \caption{GAE on CartPole using Z-score normalization. Convergence is slow with high variance.}
    \label{fig:cartpole_gae_zscore}
\end{figure}

Figure~\ref{fig:cartpole_gae_zscore} depicts training results using Generalized Advantage Estimation (GAE) with conventional normalization. Similar to the Actor-Critic case, the agent eventually converges but suffers from unstable learning in the initial episodes.

\begin{figure}[H]
    \centering
    \includegraphics[width=0.8\linewidth]{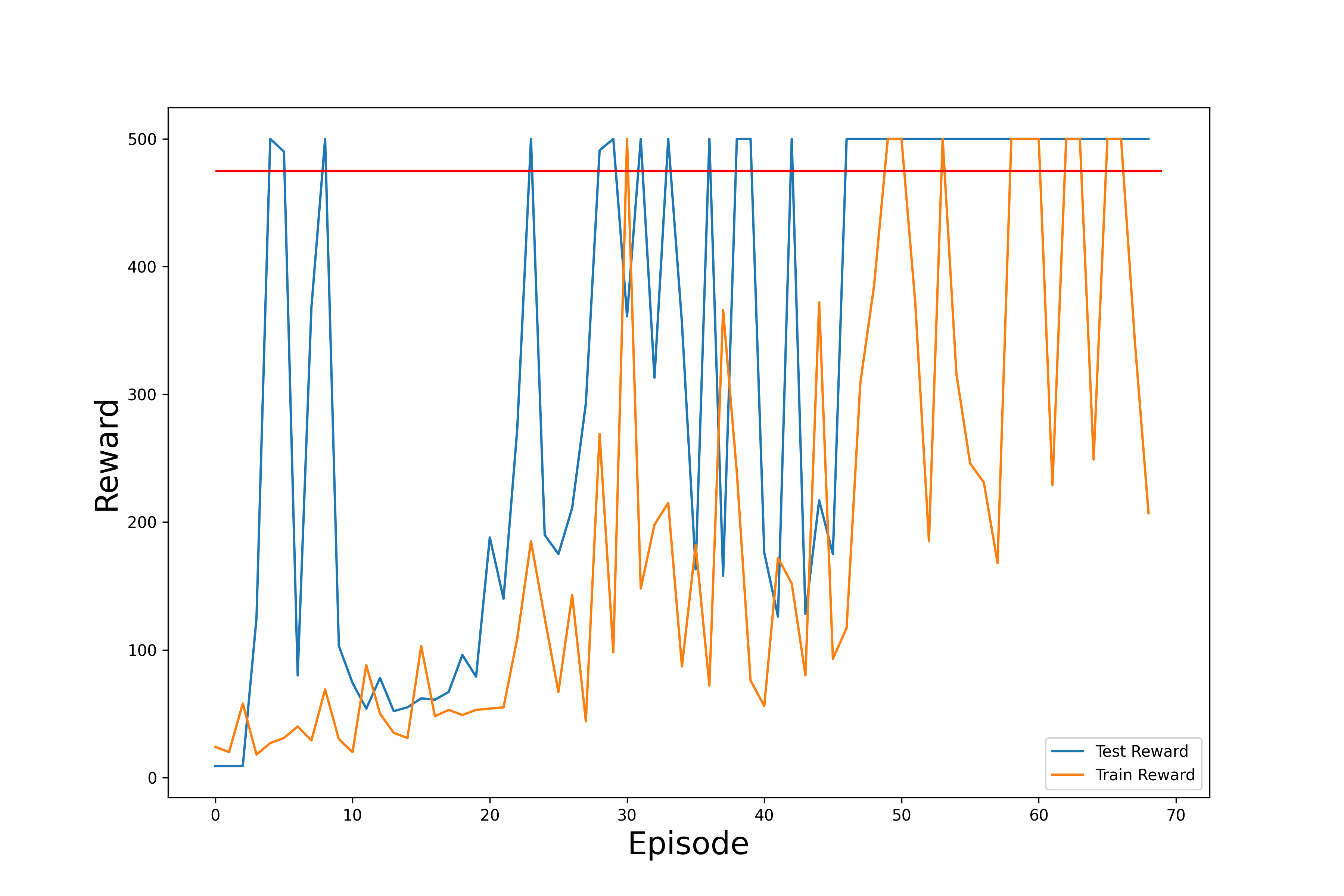}
    \caption{GAE on CartPole using Kalman normalization. Noticeably improved learning dynamics.}
    \label{fig:cartpole_gae_kalman}
\end{figure}

In contrast, Kalman-based normalization (Figure~\ref{fig:cartpole_gae_kalman}) leads to more consistent and rapid performance improvements, with less variance and more stable policy updates.

\paragraph{LunarLander with Actor-Critic}

\begin{figure}[H]
    \centering
    \includegraphics[width=0.8\linewidth]{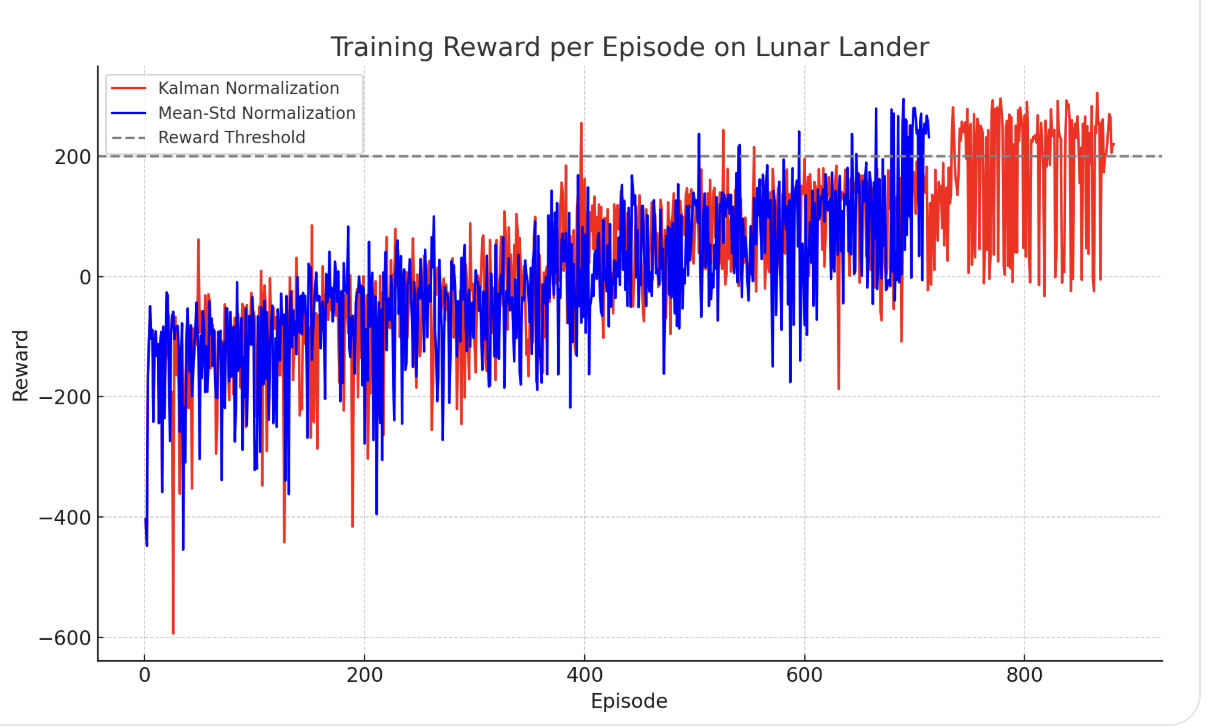}
    \caption{Actor-Critic on LunarLander with Kalman normalization. Achieves faster convergence and lower variance.}
    \label{fig:lunarlander_ac}
\end{figure}

Figure~\ref{fig:lunarlander_ac} illustrates the training rewards for the LunarLander environment using Actor-Critic with Kalman-based normalization. This environment features sparse and high-variance rewards, making stabilization particularly challenging.

Our Kalman-filtered variant outperforms mean-std normalization by achieving faster convergence and significantly smoother training curves, indicating superior handling of reward noise and non-stationarity.

\paragraph{PPO on CartPole with Multiple Normalization Methods}

To evaluate the generality of our method, we apply PPO~\cite{schulman2017proximal} with three normalization variants: adaptive Kalman, simple Kalman, and Z-score.

\begin{figure}[H]
    \centering
    \includegraphics[width=0.8\linewidth]{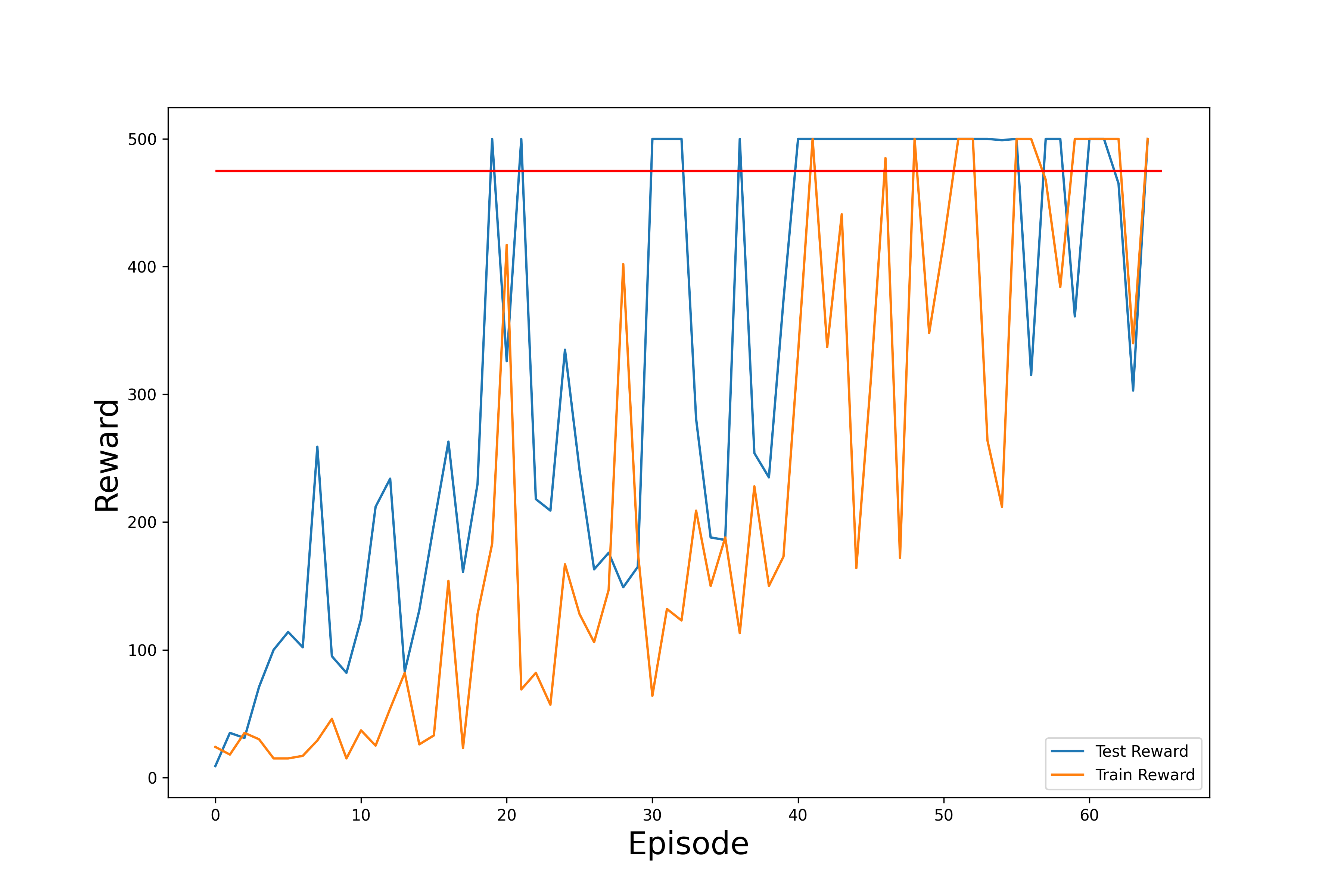}
    \caption{PPO on CartPole with Adaptive Kalman normalization. Most stable and fastest convergence.}
    \label{fig:ppo_kalman_adaptive}
\end{figure}

Figure~\ref{fig:ppo_kalman_adaptive} shows that adaptive Kalman normalization achieves the fastest convergence with minimal variance. The dynamic adjustment of measurement noise $R_t$ allows the agent to quickly adapt to changing reward patterns.

\begin{figure}[H]
    \centering
    \includegraphics[width=0.8\linewidth]{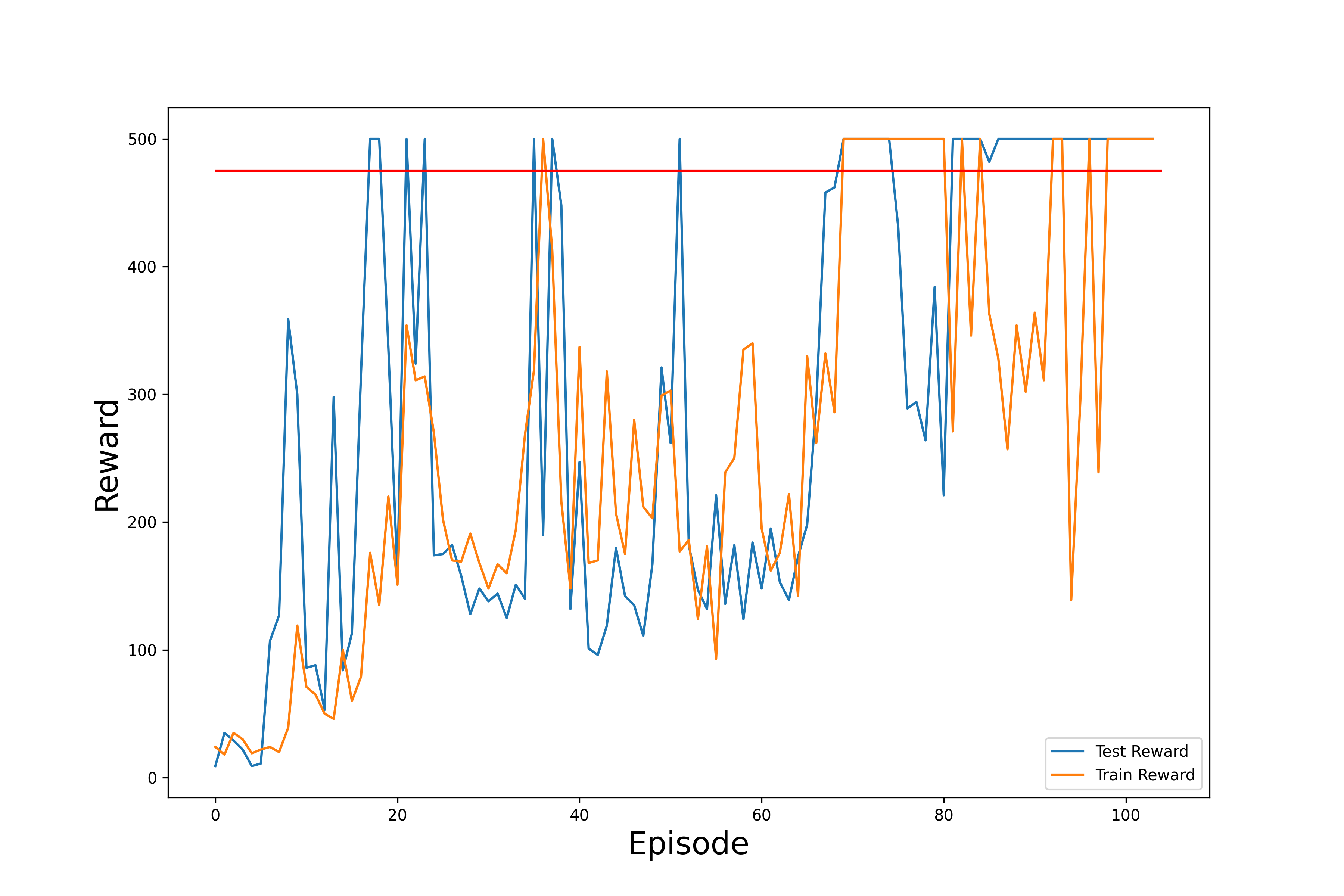}
    \caption{PPO on CartPole using Z-score normalization. Slower and noisier convergence.}
    \label{fig:ppo_zscore}
\end{figure}

Figure~\ref{fig:ppo_zscore} shows the baseline PPO performance with conventional normalization. The learning curve is noisier, and the convergence takes considerably longer.

\begin{figure}[H]
    \centering
    \includegraphics[width=0.8\linewidth]{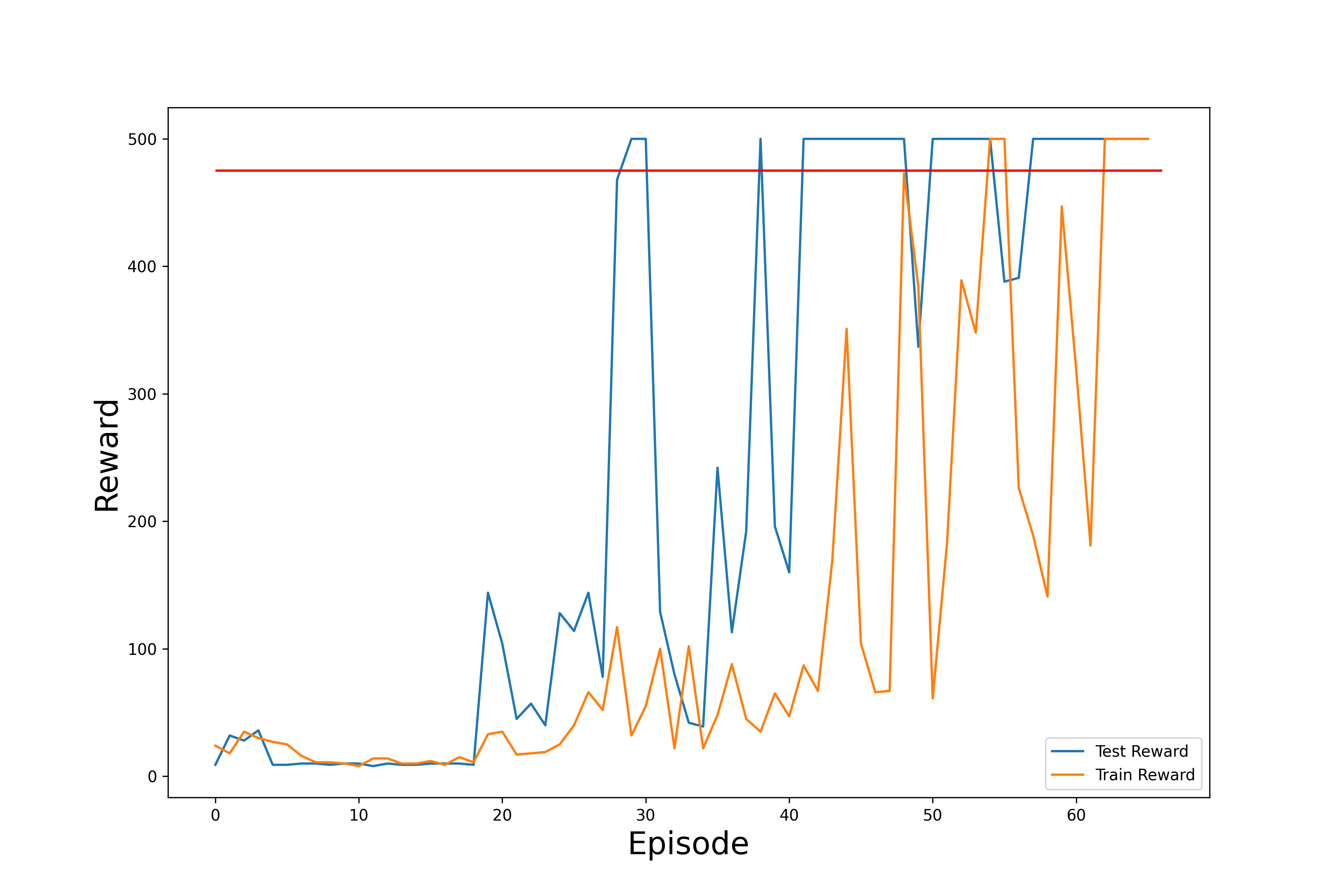}
    \caption{PPO on CartPole with Simple Kalman normalization. Intermediate performance.}
    \label{fig:ppo_kalman_fixed}
\end{figure}

Figure~\ref{fig:ppo_kalman_fixed} displays the result of using a simple Kalman filter with fixed noise parameters. While it improves over Z-score, it underperforms compared to the adaptive Kalman variant, highlighting the benefit of dynamic noise estimation.

\paragraph{Ablation Study: Effect of Process Noise $Q$}

We further study the effect of the process noise $Q$ in Kalman filtering by fixing $R = 1$ and varying $Q$ across multiple orders of magnitude. Table~\ref{tab:ablation_results} summarizes the number of episodes needed to reach the reward threshold in PPO on CartPole.

\begin{table}[H]
\centering
\caption{Effect of process noise $Q$ on convergence speed (PPO on CartPole, $R=1$ fixed)}
\label{tab:ablation_results}
\begin{tabular}{ccc}
\toprule
$Q$ & $R$ & Episodes to Threshold \\
\midrule
$1 \times 10^{-4}$ & 1 & 75 \\
$1 \times 10^{-3}$ & 1 & 71 \\
$1 \times 10^{-2}$ & 1 & 69 \\
$1 \times 10^{-1}$ & 1 & 170 \\
$1$ & 1 & 98 \\
\bottomrule
\end{tabular}
\end{table}

These results confirm that smaller values of $Q$—which imply more trust in the prior estimate—tend to accelerate convergence, as they prevent overreaction to noisy returns. However, overly small $Q$ values may lead to under-adaptation in highly non-stationary environments, suggesting a trade-off that can be mitigated by adaptive noise tuning.

\paragraph{Summary}

Across all experiments, Kalman-based reward normalization leads to faster convergence and more stable learning compared to conventional normalization. The benefits are particularly evident in early training and in environments with high reward variance. The adaptive variant consistently outperforms both fixed Kalman and Z-score baselines, validating the importance of dynamic uncertainty modeling in reinforcement learning.

\section{Conclusions, Limitations and Future Work}

\paragraph{Conclusions.}
We proposed a lightweight, principled alternative to reward normalization in policy gradient reinforcement learning based on Kalman filtering. By estimating a smoothed and adaptive baseline from the return signal, our method improves convergence speed and stability across multiple algorithms and environments. Empirical results on CartPole and LunarLander show that Kalman-based normalization outperforms standard mean-std normalization in early training and high-variance scenarios, with particularly strong gains in sample efficiency.

\paragraph{Limitations.}
Despite these promising results, our approach has several limitations. First, both the simple and adaptive Kalman variants require manual initialization of the process and measurement noise parameters $(Q, R)$. Although these can be tuned per task, the absence of an automated or learned mechanism restricts general applicability. Second, our method uses a one-dimensional Kalman filter applied solely to the scalar return signal $G_t$. While effective in simple environments, this limits the capacity to capture temporal or structural dependencies present in more complex tasks.

\paragraph{Future Work.}
Future research can explore multi-dimensional or structured extensions of the Kalman filter, enabling richer representations of uncertainty that consider state dynamics or policy structure. One natural direction is to make the Kalman filter fully differentiable and learnable within an end-to-end training framework, allowing adaptive estimation of $(Q, R)$ directly from data. Additionally, applying Kalman filtering beyond reward normalization—such as to value estimation or advantage calculation—could further improve learning stability. Finally, evaluating this approach in continuous control, offline RL, and safety-critical domains would validate its robustness and broader utility.

\section*{Accessibility}

We have taken multiple steps to improve the accessibility of this submission. All figures use colorblind-friendly color palettes and include distinct markers for different curves where appropriate. Each figure is accompanied by a descriptive caption that summarizes the main findings. We use standard LaTeX math formatting with consistent notation and avoid excessive inline equations for improved screen reader compatibility. No color is used as the sole means to convey information. The language is written in a clear and concise manner, minimizing jargon where possible. We have also ensured that the PDF passes basic accessibility checks (e.g., embedded fonts, tagged structure where applicable).

\section*{Software and Data}

To ensure reproducibility, we plan to release the full implementation of our Kalman-based reward normalization framework upon acceptance. This includes training scripts, experiment configurations, and plotting utilities used to generate all figures in the paper. The code is implemented in Python using PyTorch and will be made available as an anonymized GitHub repository or through the Supplementary Material section during the review process. All experiments are conducted on publicly available environments from OpenAI Gym (\texttt{CartPole-v1} and \texttt{LunarLander-v2}), and no proprietary data or resources are required to reproduce our results.

\section*{Impact Statement}

This paper presents a theoretically grounded method that enhances the stability and efficiency of reinforcement learning (RL) training by replacing heuristic reward normalization with a principled Kalman filtering approach. Our method is lightweight, broadly applicable, and does not modify existing network architectures, making it practical for a wide range of RL scenarios.

While the primary goal is to advance the field of machine learning, especially in improving sample efficiency and learning stability in noisy or dynamic environments, the potential societal consequences are generally aligned with those of broader RL research. For example, improved stability in RL could accelerate progress in robotics, autonomous systems, and intelligent decision-making under uncertainty.

However, as with many RL advances, care should be taken when deploying such systems in safety-critical or ethically sensitive domains. Increased learning speed and adaptability might inadvertently lead to unforeseen behaviors if not properly constrained or interpreted.

We do not foresee any specific adverse societal consequences from this work. Nevertheless, we encourage practitioners to consider the broader implications of reinforcement learning applications, particularly in areas such as surveillance, automation, and decision-making systems.

\nocite{langley00}

\bibliography{example_paper}
\bibliographystyle{icml2025}

\end{document}